\documentclass[sigconf]{acmart}
\usepackage{balance}
\usepackage{soul}
\usepackage{url}
\usepackage{wrapfig}
\usepackage{subfigure}

\def\model{GRFG}
\AtBeginDocument{%
  }
\copyrightyear{2024}
\acmYear{2024}
\setcopyright{acmlicensed}\acmConference[CIKM '24]{Proceedings of the 33rd ACM International Conference on Information and Knowledge Management}{October 21--25, 2024}{Boise, ID, USA}
\acmBooktitle{Proceedings of the 33rd ACM International Conference on Information and Knowledge Management (CIKM '24), October 21--25, 2024, Boise, ID, USA}
\acmDOI{10.1145/3627673.3680105}
\acmISBN{979-8-4007-0436-9/24/10}
\begin{document}

\title{Reinforcement Feature Transformation for Polymer Property Performance Prediction}

\author{Xuanming Hu}
\email{solomonhxm@asu.edu}
\orcid{0009-0002-2215-3553}
\affiliation{%
  \institution{Arizona State University}
  \city{Tempe}
  \state{Arizona}
  \country{USA}
}

\author{Dongjie Wang}
\email{wangdongjie@ku.edu}
\orcid{0000-0003-3948-0059}
\affiliation{%
  \institution{University of Kansas}
  \city{Lawrence}
  \state{Kansas}
  \country{USA}
}

\author{Wangyang Ying}
\email{wangyang.ying@asu.edu}
\orcid{0009-0009-6196-0287}
\affiliation{%
  \institution{Arizona State University}
  \city{Tempe}
  \state{Arizona}
  \country{USA}
}

\author{Yanjie Fu\textsuperscript{\textdagger}}
\email{yanjie.fu@asu.edu}
\orcid{0000-0002-1767-8024}
\affiliation{%
  \institution{Arizona State University}
  \city{Tempe}
  \state{Arizona}
  \country{USA}
}
\thanks{\textsuperscript{\textdagger} Corresponding Author}






\begin{abstract}
    Polymer property performance prediction aims to forecast specific features or attributes of polymers, which has become an efficient approach to measuring their performance. 
    However, existing machine learning models face challenges in effectively learning polymer representations due to low-quality polymer datasets, which consequently impact their overall performance. This study focuses on improving polymer property performance prediction tasks by reconstructing an optimal and explainable descriptor representation space. Nevertheless, prior research such as feature engineering and representation learning can only partially solve this task since they are either labor-incentive or unexplainable. This raises two issues: 1) automatic transformation and 2) explainable enhancement. To tackle these issues, we propose our unique Traceable Group-wise Reinforcement 
    Generation Perspective. Specifically, we redefine the reconstruction of the representation space into an interactive process, combining nested generation and selection. Generation creates meaningful descriptors, and selection eliminates redundancies to control descriptor sizes. Our approach employs cascading reinforcement learning with three Markov Decision Processes, automating descriptor and operation selection, and descriptor crossing. We utilize a group-wise generation strategy to explore and enhance reward signals for cascading agents.
    % employ machine learning models using physical descriptors for efficient and cost-effective polymer property performance prediction. However, the performances of these models are hindered by low-quality datasets. To tackcle this challenge, we leverage a feature transformation perspective and propose a group-wise reinforcement generation framework to enhance the accuracy of properties prediction.
    % Specifically, 
    Ultimately, we conduct experiments to indicate the effectiveness of our proposed framework.  
\end{abstract}




\maketitle
\vspace{-3mm}
\section{Introduction}
The effective design of polymers is crucial for the material industry since polymers are ubiquitously employed in lots of contexts. For instance, using polymers with high thermal conductivity is crucial for addressing heat dissipation in downsized organic devices with increasing power density~\cite{huang2023exploring}. 
Conventionally, researchers rely on costly and time-consuming experiments or simulations to verify the performance of polymers~\cite{xu2023transpolymer}.
To address it, there is a shift towards finding more efficient and cost-effective alternatives.
Researchers aim to use the physical descriptors of polymer to estimate polymer performance. This can be regarded as a regression task to predict polymer properties (i.e., polymer property performance prediction).

\begin{figure}[!t]
% \vspace{-0.2cm}
\centering
\includegraphics[width=\linewidth]{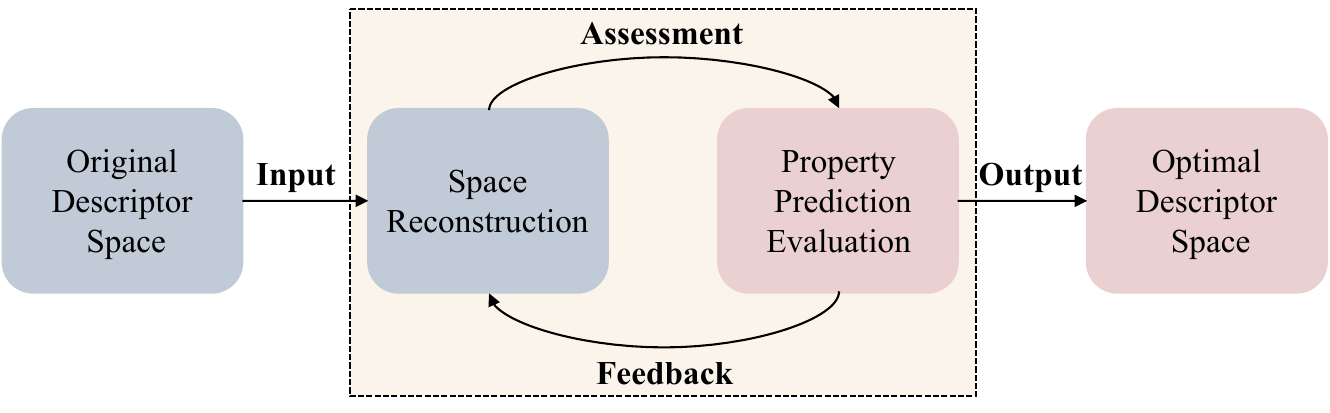}
\vspace{-5mm}
\caption{An illustration of automated transformation. We identify the optimal descriptor space that
is explicit and interpretable and performs optimally in polymer property performance prediction by iteratively reconstructing the descriptor space.}
\vspace{-0.75cm}
\label{motivation}
\end{figure}

% Machine learning NB -> Use -> hindered by low quality dataset-> feature transformation enhance
With the emergence of machine learning models in various domains for prediction~\cite{hu2023boosting,wu2020connecting}, researchers are now applying these models to polymer property performance prediction tasks. For instance, ~\cite{nagasawa2018computer} employ random forest to identify the performance of the conjugated polymers for organic solar cell. However, low-quality polymer datasets hinder existing models from effectively learning polymer representations, impacting their overall performance~\cite{huang2023exploring}.
Addressing this issue by designing more complex machine learning models can be challenging. Consequently, in this work, we forgo the approach of creating more complex models. Instead, we focus on improving polymer property performance prediction tasks by reconstructing an optimal and explainable descriptor representation space (Figure \ref{motivation}). Formally, given a set of descriptors, a specific property, the goal is to automatically generate an interpretable set of descriptors that can optimize the performance of polymer property performance prediction.
% The success of feature transformation reveals a potential solution to address the data imperfections in polymers and enhance the performance of polymer property performance prediction.
% Instead of directly applying machine learning to model the correlation between descriptors and target property, 
% Feature transformation can extract
% Specifically, we can first leverage feature transformation method to reconstruct a more discriminative space by applying mathematical operation to original descriptor (e.g., [$f_1$,$ f_2$]$\rightarrow$[$\frac{f_1}{f_2}$, $f_1-f_2$, sin($f_1$)]). Then the downstream ML models can predict the property more effectively based on the transformed descriptor space.

% Previous feature transformation shit ->Our solution
Prior works can only partially tackle this problem. Current feature engineering methods are designed to obtain a transformed data representation by preprocessing,
feature extraction, selection, and generation~\cite{guyon2003introduction,khurana2018feature,li2017feature}. But they are labor-incentive and cannot automatically generate a representation space for descriptors. Existing representation learning methods, such as factorization~\cite{fusi2018probabilistic}, embedding~\cite{goyal2018graph}, and deep representation learning~\cite{wang2021automated,wang2021reinforced}, have been wildly used to extract meaningful latent representation. The lack of explicitness and interpretability in latent representation hinders the deployment of these approaches in polymer property performance prediction, where both high predictive accuracy and a reliable understanding of performance drivers are essential.
These problems raises two key issues:
1) \textbf{Issue 1 (automatic transformation):} how can we automatically transform the descriptors to enhance polymer property performance prediction?
2) \textbf{Issue 2 (explainable enhancement):} how can we guarantee the reconstructed descriptors space is explainable?
Our objective is to provide a novel perspective in addressing these issues.

\noindent{\textbf{Our Insights: A Traceable Group-wise Reinforcement Generation Perspective.}}
We design a novel framework to tackle issues related to automation and explicitness in extracting distinctive representation space for descriptors inspired by~\cite{wang2022group}. We approach feature generation and selection through the perspective of Reinforcement Learning (RL). We demonstrate that the process of learning to reconstruct representation space can be achieved through an interactive nested generation and selection. Generation aims to create meaningful and explicit descriptors which can enhance the prediction. And selection targets to eliminate redundant descriptors that is irrelevant to predict the performance of polymers to constraint the size of descriptors. We emphasize that domain knowledge of material science can be translated into machine-learnable policies. RL enables the automatic generation of experience data and the learning of optimized policies, attracting considerable interest in recent years. Our study illustrates that the iterative sequential generation and selection can be framed as an RL task. We find that crossing descriptors with high information distinctness is more likely to generate meaningful variables in a new representation space, and thus, employing group-group crossing can enhance learning efficiency.

\noindent{\textbf{Summary of Proposed Solution.}}
We introduce a comprehensive framework, group-wise reinforcement descriptor generation, for optimal and explainable representation descriptor space reconstruction. This framework aims to achieve three main goals: 1) Goal 1: explainable explicitness, providing a traceable generation process and understanding the meanings of each generated descriptors; 2) Goal 2: self-optimization, automatically generating an optimal descriptor set for polymer property performance prediction without the involvement of domain knowledge of material science; 3) Goal 3: enhanced efficiency and reward augmentation, boosting the generation and exploration speed in a large descriptor space and augmenting reward incentive signals to learn policies.

To fulfill Goal 1, we propose an iterative descriptor generation and selection strategy. Specifically, the generation step is to mathematically transform one or two descriptors to create a new descriptor (e.g., [$f_1$,$ f_2$]$\rightarrow$[$\frac{f_1}{f_2}$, $f_1-f_2$, sin($f_1$)]) and the selection step is to control the descriptor set size.
This enables us to explicitly track the generation process and extracting semantic labels of generated descriptors. For Goal 2, we decompose descriptor generation into three Markov Decision Processes (MDPs) and develop a new cascading agent structure to coordinate agents for better selection policies. To achieve Goal 3, we design a group-operation-group-based generation approach, clustering original descriptors into different descriptor groups and letting agents select and cross two descriptor groups to generate multiple descriptors simultaneously. This strategy accelerates representation space reconstruction by exploring the descriptor space faster and addressing issues related to reward signal insufficiency.

% \noindent{\textbf{Our contributions}} are: 1)
% 2)
% 3)
\vspace{-2mm}
\section{Related Works}
\noindent{\textbf{Polymer Property Performance Prediction}} aims to forecast specific characteristics or behaviors of polymers to estimate their performance. As machine learning models have become prevalent in diverse domains for prediction, earlier studies also utilized these models for predictive tasks~\cite{karamad2020orbital,jha2019enhancing}. However, the inadequate quality of polymer datasets poses a challenge for current models, preventing them from acquiring polymer representations and thereby affecting their overall performance.
In this paper, we improve polymer property performance prediction tasks by reconstructing an optimal and explainable descriptor representation space.

\noindent{\textbf{Reinforcement Learning (RL)}}
involves determining how agents should response in a dynamic environment to maximize rewards~\cite{sutton2018reinforcement}. RL algorithms can be categorized as value-based or policy-based, depending on whether they estimate the value of states~\cite{mnih2013playing,van2016deep} or learn a probability distribution for action selection~\cite{sutton1999policy}. An actor-critic RL framework is introduced to combine the strengths of both approaches~\cite{schulman2017proximal}. In this paper, we model the selection of descriptor groups and operations as Markov Decision Processes (MDPs) and propose a cascading agent structure to solve these MDPs.

\noindent{\textbf{Feature Engineering}}
seeks to enhance the feature space through generation and selection, aiming to improve machine learning model performance~\cite{zheng2018feature}. Feature selection involves removing redundant features~\cite{ying2024feature,gong2024neuro,wang2024knockoff}, while feature generation creates meaningful variables~\cite{ying2023self,ying2024unsupervised}. Selection methods include filter~\cite{liu1996probabilistic}, wrapper~\cite{liu2021efficient}, and embedded approaches~\cite{tibshirani1996regression}, while generation methods involve latent representation learning~\cite{blei2003latent} and feature transformation~\cite{chen2019neural,wang2024reinforcement,xiao2023traceable,xiao2024traceable,huang2024enhancing,xiao2023beyond}. Unlike existing methods, our personalized descriptor crossing, cascading agents, and group-wise generation strategy are designed for polymer property
performance prediction and are able to address descriptor distinctness, accelerate generation, and capture interactions.

% \vspace{-3mm}
\section{Preliminaries}
\begin{figure*}[!t]
% \vspace{-0.2cm}
\centering
\includegraphics[width=0.85\linewidth]{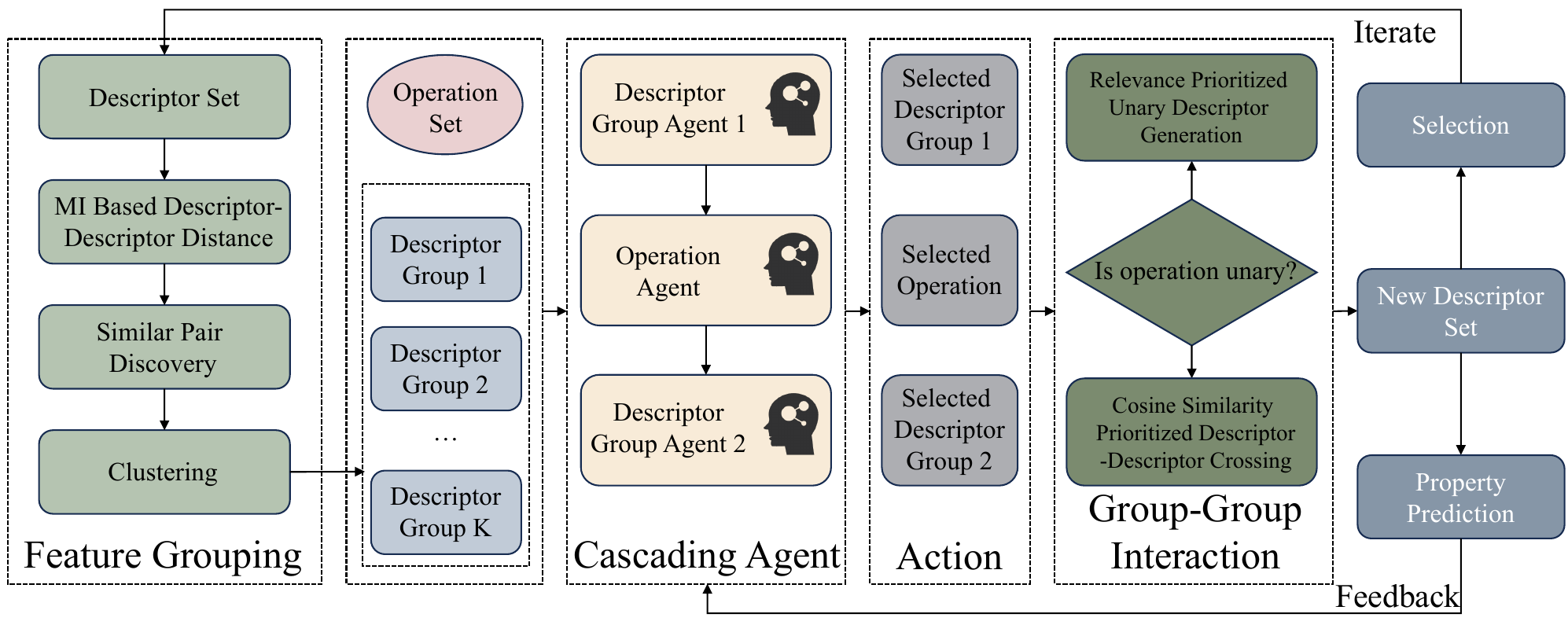}
% \vspace{-0.5cm}
\caption{An overview of our proposed framework.}
% \vspace{-0.6cm}
\label{framework}
\end{figure*}
% \vspace{-1mm}
\subsection{Important Definitions}
\textit{Definition 2.1.}  \textbf{Descriptor Group.} We target to reconstruct the descriptor space of datasets $\mathcal{D}=\{\mathcal{F},y\}$, where $\mathcal{F}$ represents a descriptor set, with each column representing a feature and each row representing a data sample, and $y$ is the target property set corresponding to samples. To generate new descriptors effectively, we partition the physical descriptor set $\mathcal{F}$ into distinct physical descriptor groups through clustering, denoted by $\mathcal{C}$, with each physical descriptor group being a subset of descriptors from $\mathcal{F}$.

\noindent\textit{Definition 2.2.}  \textbf{Operation Set.} We conduct mathematical operations on current descriptors to create new ones. The set of all operations, denoted as $\mathcal{O}$, consists of two types: unary operations, such as "square," "exp," "log," and others, and binary operations, including "plus," "multiply," "divide,", etc.

\noindent\textit{Definition 2.3.}  \textbf{Cascading Agent.} To enhance the accuracy of prediction, we introduce a novel cascading agent structure consisting of three agents: two physical descriptor group agents and one operation agent. These cascading agents share state information and sequentially choose physical descriptor groups and operations.

% \vspace{-3mm}
\subsection{Problem Statement}
The research problem of this paper is to regenerate an optimal space for descriptors to improve the accuracy of polymer property performance prediction task. Formally, given a dataset $\mathcal{D}=\{\mathcal{F},y\}$ and an operator set $\mathcal{O}$, our goal is to automatically reconstruct an optimal physical descriptor set $\mathcal{F}^{\ast}$ that maximizes the performance indicator $V$ of the polymer property performance prediction task $A$, which can be formulated as:
\begin{equation}
    \mathcal{F}^{\ast}=\mathop{\arg\max}\limits_{\mathcal{\hat{F}}}(V_A(\mathcal{\hat{F}},y))
\end{equation}
where $\mathcal{\hat{F}}$ is a subset formed by combining the original descriptor set $\mathcal{F}$ and the generated new descriptors $\mathcal{F}^g$. $\mathcal{F}^g$ is created by applying the operations in $\mathcal{O}$ to the original physical descriptor set $\mathcal{F}$ using a specific algorithmic structure.

\vspace{-2mm}
\section{Methodology}
\vspace{-1mm}
\subsection{Framework Overview}
Figure~\ref{framework} presents our proposed framework, namely Group-wise Reinforcement Feature Generation (GRFG). Firstly, we maximize intra-group
feature similarity and inter-group feature distinctness to cluster original descriptors into distinct groups. Then, using a group-operation-group strategy, two descriptor groups are crossed to generate multiple descriptors simultaneously. A cascading reinforcement learning method is leveraged to guide three agents in selecting informative descriptor groups and operations cooperatively. The cascading method facilitates coordinated state sharing among agents, optimizing their choice policies. After selection, new descriptors are generated using a group-group crossing strategy. In the third step, the generated descriptors are added to the original set, forming a new set. This set is used in a polymer property performance prediction task to collect predictive accuracy as reward feedback, updating policy parameters. Finally, descriptor selection reduces redundancies, continuing iterations until the maximum limit is reached.

\vspace{-4mm}
\subsection{Descriptor Clustering for Generation}
\vspace{-1mm}
Previous automated feature engineering technique like transformation graph~\cite{khurana2018feature} and neural feature search~\cite{chen2019neural} can partially solve our task. Transformation graph applies the selected operation to the entire descriptor set without considering the heterogeneous physical characteristics among descriptors, leading to the generation of sub-optimal descriptors. In neural feature search, a recurrent neural network (RNN) is employed for each descriptor to learn its descriptor transformation sequence for descriptor generation. However, this approach neglects the distinctness among descriptors, limiting the its ability to generate meaningful combined descriptors. 

Therefore, to consider distinctive characteristics among descriptors, we leverage group-wise descriptor generation strategy to integrate the reward feedback of agents for policy learning. To generate meaningful descriptor groups suitable for subsequent group-group crossing, our framework initiates with a generation-oriented descriptor clustering approach. However, the crossing of descriptors with high (low) information distinctness is more (less) likely to generate meaningful variables in a new representation space. Consequently, unlike traditional clustering methods, our goal is to group descriptors into distinct descriptor groups, with the optimization objective of maximizing inter-group descriptor information distinctness while minimizing intra-group descriptor information distinctness. Specifically, we propose M-Clustering for clustering, which treats each descriptor as an individual group initially and subsequently merges the closest descriptor group pair at each iteration.

To accomplish the objective of minimizing intra-group descriptor difference and maximizing inter-group descriptor dissimilarity, we introduce a novel distance measure to quantify the distinctness between two descriptor groups. Two key observations are emphasized: 1) Relevance to property: If the relevance between one descriptor group and the property is comparable to the relevance of another descriptor group and the predictive target, then the two descriptor groups are considered similar. The fundamental reason is that these physical descriptor share resemble mechanism to estimate the performance of polymers. 2) Mutual information: If the mutual information between the descriptors of two groups is substantial, the two descriptor groups are considered similar. Leveraging these insights, we formulate a descriptor group-group distance. This distance serves as a metric to quantify the distinctness of two descriptor groups and provides insights into how likely crossing the two descriptor groups will result in the generation of more meaningful descriptors.
The distance is calculated by:
\begin{equation}
    dis(C_i, C_j) = \frac{1}{|C_i|\cdot|C_j|}\sum_{f_i\in C_i}\sum_{f_j\in C_j}\frac{|MI(f_i,y)-MI(f_j,y)|}{MI(f_i,f_j)+\epsilon}
\end{equation}
where $C_i$ and $C_j$ represent two different descriptor groups, $|C_i|$ and $|C_j|$ are the numbers of descriptors in $C_i$ and $C_j$ , $f_i$ and $f_j$ are two descriptors belongs to $C_i$ and $C_j$ respectively, $y$ is the target property. In
particular, $|MI(f_i,y)-MI(f_j,y)|$ measures the difference in relevance
between $y$ and $f_i$, $f_j$ . A smaller value suggests that $f_i$ and $f_j$ have a more similar influence on classifying $y$. $MI(f_i,f_j)+\epsilon$ measures the redundancy between $f_i$ and $f_j$ . $\epsilon$ is a small value that
is used to prevent the denominator from being zero. A larger value indicates that $f_i$ and $f_j$ share more information.

% The Feature Group Distance based M-Clustering Algorithm is designed to address the challenge of non-spherical feature clusters by introducing a group-group distance instead of a point-point distance. 

To encourage non-spherical feature clusters, we develop Descriptor Group Distance based M-Clustering Algorithm by introducing a group-group distance instead of a point-point distance.
This makes traditional methods like K-means or density-based methods inappropriate. Inspired by agglomerative clustering, the algorithm follows a three-step process:
1) Initialization: Each descriptor in the descriptor set $\mathcal{F}$ is considered as one single descriptor cluster.
2) Repeat: Information overlap between any two descriptor clusters is calculated, and the cluster pair with the closest overlap is identified. The two clusters are then merged into one, and the original clusters are removed.
3) Stop Criteria: The Repeat step is iterated until the distance between the closest feature group pair reaches a threshold. While the classic stop criterion is to terminal when there is only one cluster, using the distance between the closest feature groups as the stopping criterion enhances our ability to semantically understand and identify the information distinctness among feature groups. This approach also facilitates practical implementation in a deployment scenario for polymer property performance prediction.

% \vspace{-3mm}
\subsection{Selection of Descriptors and Operation}
% \vspace{-1mm}
\begin{figure}[!t]
% \vspace{-0.2cm}
\centering
\includegraphics[width=0.9\linewidth]{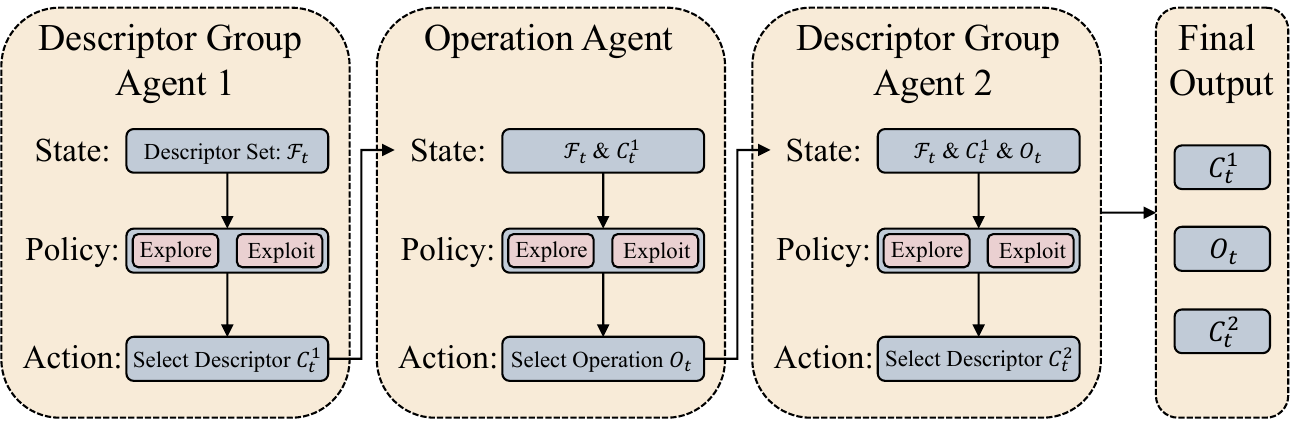}
% \vspace{-0.3cm}
\caption{The cascading agents include the descriptor group agent 1, the operation agent, and the descriptor group agent 2. They choose two candidate descriptor groups and a operation.}
% \vspace{-7mm}
\label{agent}
\end{figure}
To enable group-wise generation, we choose two descriptor group, and an operation for the combination of group-operation-group based crossing. Two critical observations drive our use of cascading reinforcement. Firstly, while it's challenging to define and program optimal selection criteria for descriptor groups and operations due to the complex chemical mechanism between descriptors and property, we can consider selection criteria as machine-learnable policies. 
To this end, we introduce three agents to learn the material knowledge implicitly via these policies through trials and errors. We discover that the three selection agents form a cascading, sequential, auto-correlated decision structure, rather than operating independently. This is because selected descriptors and operations are highly correlated, reflecting implicit material knowledge. In this context, "cascading" means that during each iteration, agents make decisions sequentially, with downstream agents awaiting the outcomes of upstream agents. The decision of an upstream agent alters the environmental states for downstream agents.

As depicted in Figure~\ref{agent}, the first agent selects the initial descriptor group based on the overall descriptor space state. The second agent then chooses the operation, considering the entire descriptor space and the selection made by the first agent. Finally, the third agent selects the second descriptor group based on the overall descriptor space and the choices made by the first and second agents. We subsequently introduce two generic metrics to quantify the usefulness (reward) of a descriptor set and propose three Markov Decision Processes (MDPs) to learn the policies of the three selection agents.

\noindent{\textbf{Metric 1: Combined Descriptor Set Redundancy and Relevance (Unsupervised Perspective).}}
We introduce a metric to assess the utility of a descriptor set from an information theory perspective. A descriptor set with higher utility contains less redundant information and is more relevant to the performance of polymers. Formally, for a given descriptor set $\mathcal{F}$ and a target property label $y$, the calculation process is
\begin{equation}
U(\mathcal{F}|y)= -\frac{1}{|\mathcal{F}|^2}\sum_{f_i,f_j\in \mathcal{F}}MI(f_i,f_j)+\frac{1}{|\mathcal{F}|}\sum_{f\in \mathcal{F}}MI(f,y)
\end{equation}

where $MI$ represent the mutual information, $f_i$, $f_j$, and $f$ are descriptor in $\mathcal{F}$, and $|\mathcal{F}|$ is the corresponding size.

\noindent{\textbf{Metric 2: polymer property performance prediction Accuracy (Supervised Perspective).}} Since our task is to enhance the performance of polymer property performance prediction, we need one more utility metric to measure whether our learned representation space can improve the accuracy of prediction. Here, we utilize 1-RAE as the performance indicator for evaluation.
Utilizing the two metrics, we proceed to create three Markov Decision Processes (MDPs) to train three agent policies for selecting the optimal feature group 1, operation, and feature group 2.

\noindent{\textbf{Selection Agent of Descriptor Group 1.}}
The descriptor group 1 agent is designed to select the optimal meta descriptor group 1. Its learning system for t-th iteration includes: 
1) Action: the meta descriptor group 1 selected from the descriptor groups of the previous iteration, denoted as $a_t^1=C_{t-1}^1$.
2) State: vectorized representation of the generated descriptor set of the previous iteration, denoted as $s_t^1=Rep(\mathcal{F}_{t-1})$, where $Rep(\cdot)$ is a state representation method which we will discussed later.
3) Reward: utility score of the selected descriptor group 1, denoted as $\mathcal{R}(s_t^1, a_t^1)= U(\mathcal{F}_{t-1}|y)$.

\noindent{\textbf{Selection Agent of Operation.}}
The operation agent is designed to select the optimal operation from an operation set to generate new descriptor. Its learning system includes: 
1) Action: selected operation, denoted as $a_t^o = o_t$.
2) State: the combination of $Rep(\mathcal{F}_{t-1})$ and the representation
of the descriptor group selected by the previous descriptor group 1 agent, denoted as $s_t^o=Rep(\mathcal{F}_{t-1})\oplus Rep(C_{t-1}^1)$, where $\oplus$ is the concatenation operation.
3) Reward: The selected operation is employed to generate new descriptors through descriptor crossing. These new descriptors are then combined with the original descriptor set to create a new descriptor set. Therefore, the updated descriptor set at the t-th iteration can be written as $\mathcal{F}_t= \mathcal{F}_{t-1}\cup g_t$, where $g_t$ is the new generated descriptors. The reward is the improvement in the utility score of the descriptor set,
denoted as $\mathcal{R}(s_t^o, a_t^o)= U(\mathcal{F}_{t}|y)- U(\mathcal{F}_{t-1}|y)$.

\noindent{\textbf{Selection Agent of Descriptor Group 2.}}
The descriptor group 2 agent is to select the optimal meta descriptor group 2. Its learning system includes: 
1) Action: the meta descriptor group 1 selected from the descriptor groups of the previous iteration, denoted as $a_t^2=C_{t-1}^2$.
2) State: the combination of $Rep(\mathcal{F}_{t-1})$, $Rep(C_{t-1}^1)$ and the representation of the operation selected, denoted as $s_t^2=Rep(\mathcal{F}_{t-1})\oplus Rep(C_{t-1}^1)\oplus Rep(o_t)$,
3) Reward: the improvement of descriptor set utility and polymer property performance prediction, denoted as $\mathcal{R}(s_t^2, a_t^2)= U(\mathcal{F}_{t}|y)- U(\mathcal{F}_{t-1}|y) + V_{A_t}$, where $V_A$ represents the performance of polymer property prediction.

\noindent{\textbf{State Representation.}}
We propose a explicit descriptor group representation method for creating a vector characterizing the state of the descriptor group. Firstly, for a descriptor group $\mathcal{F}$, we calculate column-wise descriptive statistics (count, standard deviation, minimum, maximum, first, second, and third quartile). Subsequently, row by row, we compute descriptive statistics for the previous step's outcome, resulting in a descriptive matrix of shape 
$\mathbb{R}^{7\times7}$. Flattening this matrix yields the feature group's representation 
$Rep(\mathcal{F})\in \mathbb{R}^{1\times49}$. This representation method generates a fixed-size state vector, accommodating variations in descriptor set size across generation iterations. Secondly, for the operation, we employ one-hot encoding as its representation $Rep(o)$.

\noindent{\textbf{Optimization Process.}}
The optimization objective of these three agents is to maximize the discounted and cumulative reward throughout the iterative descriptor generation process. We incentivize the cascading agents to work cooperatively to produce a descriptor set that is informative and effective in the downstream task. To achieve this objective, we minimize the temporal difference error $\mathcal{L}$, which is derived from the Bellman equation and is expressed as
\begin{equation}
    \mathcal{L} = Q(s_t, a_t)-(\mathcal{R}(s_t, a_t)+\gamma max_{a_{t+1}}Q(s_{t+1}, a_{t+1}))
\end{equation}
Here, $\gamma\in[0\sim 1]$ is the discounted factor; $Q$ is the $Q$ function approximated by deep neural networks. After the convergence of the agents, we anticipate to identify the optimal policy $\pi^{\ast}$ that can choose the most appropriate descriptor group or operation based on the
state via the estimated Q-value, which can be calculated as:
\begin{equation}
    \pi^{\ast}(a_t|s_t)= argmax_{a}Q(s_t, a)
\end{equation}

% \vspace{-4mm}
\subsection{Group-wise Descriptor Generation}
% \vspace{-1mm}

\noindent{\textbf{Scenario 1: Similarity Based Descriptor-Descriptor Crossing.}}
We emphasize the likelihood of generating informative descriptors by crossing two descriptors that exhibit lower overlap in terms of information. We propose a simple yet effective approach, i.e., selecting the top K dissimilar descriptor pairs between two descriptor groups. Specifically, this involves crossing the groups, calculating cosine similarities for all descriptor pairs, ranking them, and choosing the top K dissimilar pairs. The selected pairs then undergo a specified operation, generating K new descriptors.

\noindent{\textbf{Scenario 2: Relevance Based Unary Descriptor Generation.}}
% \subsection{polymer property performance prediction via generated Descriptors}
When an unary operation and two descriptor groups are selected, we proceed by applying the operation to the descriptor group that exhibits greater relevance to the target property. For a given descriptor group $C$, we gauge the relevance between the descriptor group and the target property by calculating the average mutual information between all the descriptors in $C$ and the property $y$. This relevance measure is denoted as $rel$ and is computed as follows:
\begin{equation}
    rel = \frac{1}{|C|}\sum_{f_i\in C}MI(f_i,y)
\end{equation}
Here, MI is mutual information. Once the more relevant descriptor group is identified, the unary operation is applied to each descriptor within that group, resulting in the generation of new descriptors.

\noindent{\textbf{Post-generation Processing.}}
Following generation, the newly created descriptors are amalgamated with the original descriptor set to construct an updated descriptor set. This updated set is then utilized in polymer property performance prediction to assess predictive performance. The performance achieved serves as reward feedback, guiding the optimization of policies for the three cascading agents in preparation for the subsequent round of descriptor generation. To mitigate descriptor number explosion throughout the iterative generation process, a selection step is implemented to regulate descriptor size. When the size of the new descriptor set surpasses a predefined size tolerance threshold, the K-best selection method is employed to reduce the descriptor size. Otherwise, selection is not performed. The new descriptor set subsequently becomes the original descriptor set for the subsequent iteration. Ultimately, when the framework is convergent, the optimal descriptor set $\mathcal{F}^{\ast}$ which has the highest accuracy for polymer property performance prediction will be returned.
% \vspace{-5mm}
\section{Experiments}
% \vspace{-1mm}
\subsection{Experimental Setup}
% \vspace{-1mm}
\noindent{\textbf{Data Description.}}
We utilize the polymers dataset generated by~\cite{huang2023exploring}, where the target property is thermal conductivity, and the number of physical descriptors has been narrowed down to 20.

\noindent{\textbf{Evaluation Metrics.}}
For fair comparison, we employ Random Forests (RF) as a downstream model for polymer property performance prediction to validate the effectiveness of learned representation space due to its stability. 
To measure the accuracy of polymer property performance prediction, we use 1-Relative Absolute Error (1-RAE), 1-Mean Average Error (1-MAE), and 1-Mean Square Error (1-MSE). Specifically,
    $1-\mathrm{RAE} = 1-\frac{\sum_{i=1}^n|y_i-\hat{y}_i|}{\sum_{i=1}^n|y_i-\overline{y}_i|}$,
$1-MAE = 1-\sum_{i=1}^n|y_i-\hat{y}_i|$,
$1-MSE = 1-\sum_{i=1}^n(y_i-\hat{y}_i)^2$, where $n$ is the number of data points, $y_i$, $\hat{y}_i$, $\overline{y}_i$ are the ground-truth property values, predicted property values, and the mean of ground-truth property, respectively.

\noindent{\textbf{Baseline Algorithm.}}
We compare our framework with six wildly-used feature transformation methods: 1) \textbf{Original (ORG)} represent the descriptor space without transformation;
2) \textbf{Random Generation (RDG)} randomly generates descriptor-operation-descriptor transformations to construct a new descriptor space; 3) \textbf{Essential Random Generation (ERG)} initially expands the descriptor space by applying operations to each descriptor, and then chooses the essential descriptors as the new descriptor space; 4) \textbf{Latent Dirichlet Allocation (LDA)}~\cite{blei2003latent} refines the descriptor space to get the factorized hidden state via matrix factorization; 5) \textbf{AutoFeat Automated Transformation (AFAT)}~\cite{horn2020autofeat} is an enhanced version of ERG, which repeatedly generates new descriptors and employs multi-step descriptor selection to identify informative ones; 6) \textbf{Neural Feature Search (NFS)}~\cite{chen2019neural} generates the transformation sequence for each descriptor and the entire process is optimized by RL; 7) \textbf{Traversal Transformation Graph (TTG)}~\cite{khurana2018feature} formulates the transformation process as a graph and subsequently employs an RL-based search method to identify the optimal descriptor set.
% \vspace{-3mm}
\subsection{Experimental Result}
% \vspace{-1mm}
\subsubsection{Overall Comparison}
\setlength{\tabcolsep}{2.7mm}{
\begin{table}[tb]
\centering
\caption{Overall performance, where $\Delta_{\rm{1-RAE}}$ represents the improvement of 1-RAE compared to ORG.}
\vspace{-0.3cm}
\begin{tabular}{@{}c|cccc@{}}
\toprule \toprule
&1-MAE             & 1-MSE            & 1-RAE  
&  
$\Delta_{\rm{1-RAE}}$ \\
\midrule
ORG     & 0.467          & 0.460          & 0.183 & +0.00\%           \\
RDG     & 0.467          & \underline{0.464}          & 0.184 & +0.01\%      \\
ERG     & 0.466          & 0.450          & 0.183          & +0.00\%                  \\
LDA     &  0.334         & 0.247          & 0.022          & \textminus88.0\%                   \\
AFAT    & 0.462          & 0.460          &0.178          & \textminus0.03\%           \\
NFS     & \underline{0.468}          & 0.455          & \underline{0.185}          & +0.01\%                \\
TTG     & 0.466  & 0.456    & 0.183          & +0.00\%           \\
\midrule
\model & \textbf{0.480} & \textbf{0.469} & \textbf{0.206}    & \textbf{+12.6\%} \\
\bottomrule \bottomrule
\end{tabular}
\vspace{-0.5cm}
\label{exp:main}
\end{table}}
We compare the descriptor transformation performance of \model\ with other baseline models. Table~\ref{exp:main} indicates the
overall comparison results in terms of 1-MAE, 1-MSE and 1-RAE. We
observe that \model\ consistently outperforms other baseline models
across all evaluation metrics. The underlying driver for this observation is that personalized descriptor crossing strategy leveraged by \model\ can take the heterogeneous descriptor-descriptor into account and reconstruct meaningful descriptor for polymer property performance prediction. Moreover, Additionally, the finding that GRFG outperforms random-based (RDG) and expansion-reduction-based (ERG, AFT) methods indicates that the agents are capable of sharing states and rewards in a cascading fashion. Consequently, they can learn an effective policy for selecting optimal crossing features and operations. In addition, due to the self-learning policy of RL agents, our proposed framework can automatically extract the hidden material knowledge among physical descriptors and property of polymers. Therefore, our proposed framework can be apply to any polymer property performance prediction scenarios without the involvement of human experts.

\subsubsection{Ablation Study}
\begin{figure}[t]
% \vspace{-0.6cm}
        \centering
        \subfigure[1-MAE]{\includegraphics[width=0.155\textwidth]{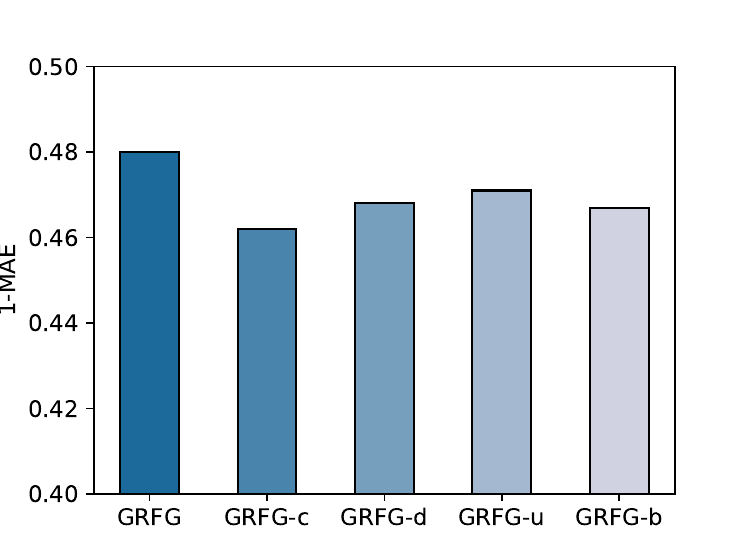}}
        \subfigure[1-MSE]{\includegraphics[width=0.155\textwidth]{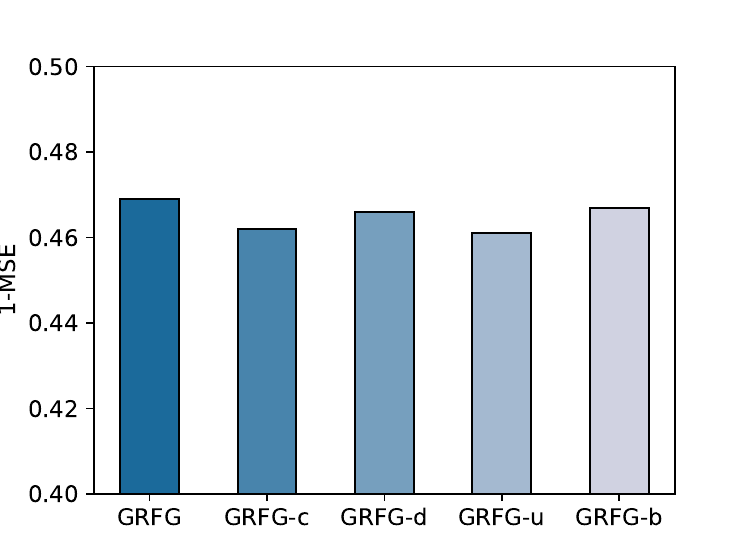}}
        \subfigure[1-RAE]{\includegraphics[width=0.155\textwidth]{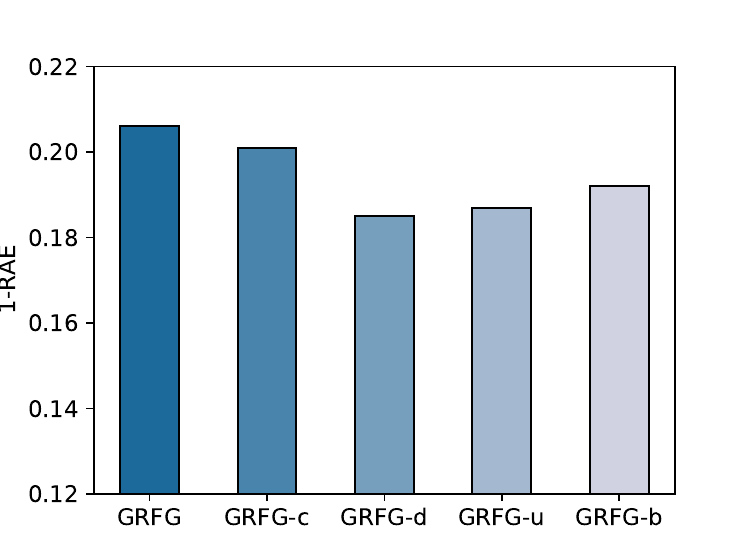}}
        % \vspace{-5mm}
        \caption{Ablation study for \model. In this experiments, we validate the necessity of different components in GRFG.}
        % \vspace{-5mm}
        \label{ablation}
\end{figure}
\setlength{\tabcolsep}{2.7mm}{
\begin{table}[tb]
\centering
\small
\caption{Robustness check on ML models. We verify whether \model\ can generate a robust descriptor space.}
% \vspace{-3mm}
\begin{tabular}{@{}c|cccccc@{}}
\toprule \toprule
& RF             & SVM            & KNN            & DT             & LASSO          & Ridge          \\
\midrule
ORG     & 0.183          & 0.113          &0.086  &  -0.102         &  -0.013  &   0.096        \\
RDG     & 0.184          &    0.126       & 0.079 &    -0.071       &  \underline{-0.012}  & \underline{0.104}         \\
ERG     & 0.183          & -0.326          & -0.065          & -0.112          & -0.015          & 0.142          \\
LDA     & 0.022        & 0.018          & -0.088          & -0.013          & -0.013          & -0.015          \\
AFAT    & 0.178          & \underline{0.131}          & 0.065          & -0.078   & -0.018          & 0.040          \\
NFS     & \underline{0.185}          &    0.112      &    \underline{0.087}      &   \underline{-0.024}       &  -0.013   &  0.083       \\
TTG     & 0.183  &  0.117   &    0.072      &    -0.095      &   -0.013    & 0.102  \\
\midrule
\model & \textbf{0.206} & \textbf{0.180} & \textbf{0.090}    & \textbf{0.020} & \textbf{0.013} & \textbf{0.151} \\
\bottomrule \bottomrule
\end{tabular}
% \vspace{-0.5cm}
\label{exp:down}
\end{table}}

To evaluate the necessity of various components in \model, we develop four model variants: 
1) \model-c removes the clustering step of \model\ and create descriptors by descriptor-operation-descriptor based crossing, instead of group-operation-group based crossing.
2) \model-d employ the euclidean distance as the metric for M-Clustering.
3) \model-u picks a descriptor group at random from the descriptor group set, when the operation is unary.
4) \model-b randomly selects descriptors from the larger descriptor group to
align two descriptor groups when the operation is binary.
Figure~\ref{ablation} demonstrates the results of the ablation study. We can observe that each component of \model\ is effective to capture hidden material knowledge and construct an optimal representation space for descriptor to enhance polymer property performance prediction.
Specifically, GRFG-c exhibits poorer performance than GRFG across all metrics. This underscores the concept that group-level generation can enhance reward feedback, aiding cascading agents in learning more effective policies.
In addition, the fact that GRFG outperforms GRFG-d indicates that our distance metric effectively captures group-level information relevance and redundancy ratio. This approach aims to maximize information distinctness across descriptor groups while minimizing it within a descriptor group. Such a distance metric assists GRFG in generating more useful dimensions.
Moreover, we notice that both GRFG-u and GRFG-b exhibit poorer performance compared to GRFG. This serves as validation that combining two distinct descriptors and prioritizing generation based on relevance can lead to the creation of superior descriptors.

\subsubsection{Robustness Check}
To check the robustness of \model\ for distinct downstream ML models, we replace the downstream model with Random Forest (RF), Support Vector Machine (SVM), K-Nearest Neighborhood (KNN), Ridge, LASSO, and Decision Tree (DT). 
Table~\ref{exp:down} shows the comparison results in terms of the 1-RAE. 
We find that \model\ outperforms baselines robustly in all cases. 
A potential reason for this observation is that \model\ can generate an optimal, robust and explainable representation space to enhance the polymer property performance prediction. Therefore, regardless of the downstream ML models, the representation space generated by \model\ can always improve the performance of polymer property performance prediction by improving data quality automatically.
This experiment demonstrates that the descriptor space generated \model\ is robust across different ML models.

\subsubsection{Hyperparameter Sensitivity}
\begin{figure}[t]
% \vspace{-1mm}
        \centering
        \subfigure[1-MAE]{\includegraphics[width=0.155\textwidth]{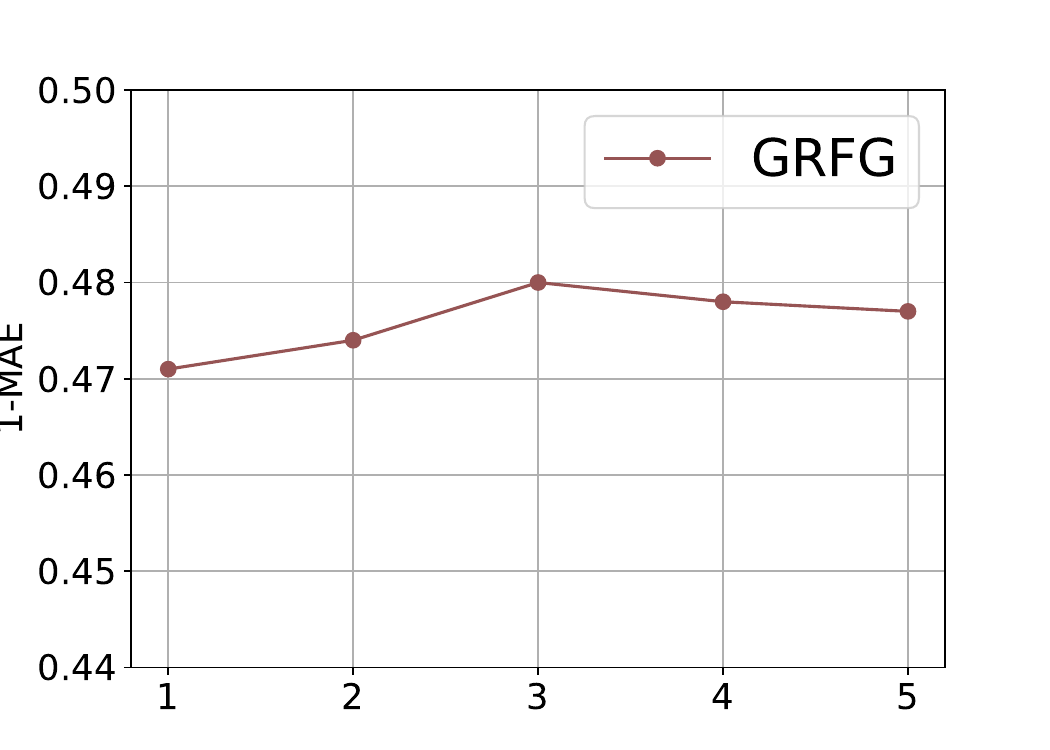}}
        \subfigure[1-MSE]{\includegraphics[width=0.155\textwidth]{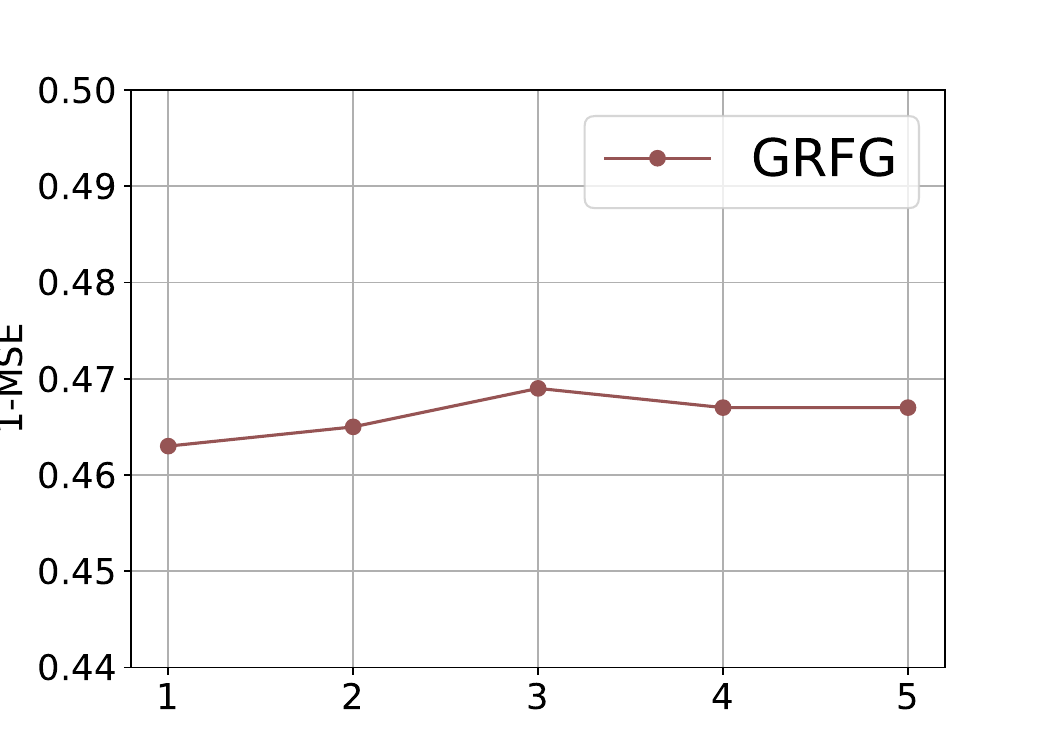}}
        \subfigure[1-RAE]{\includegraphics[width=0.155\textwidth]{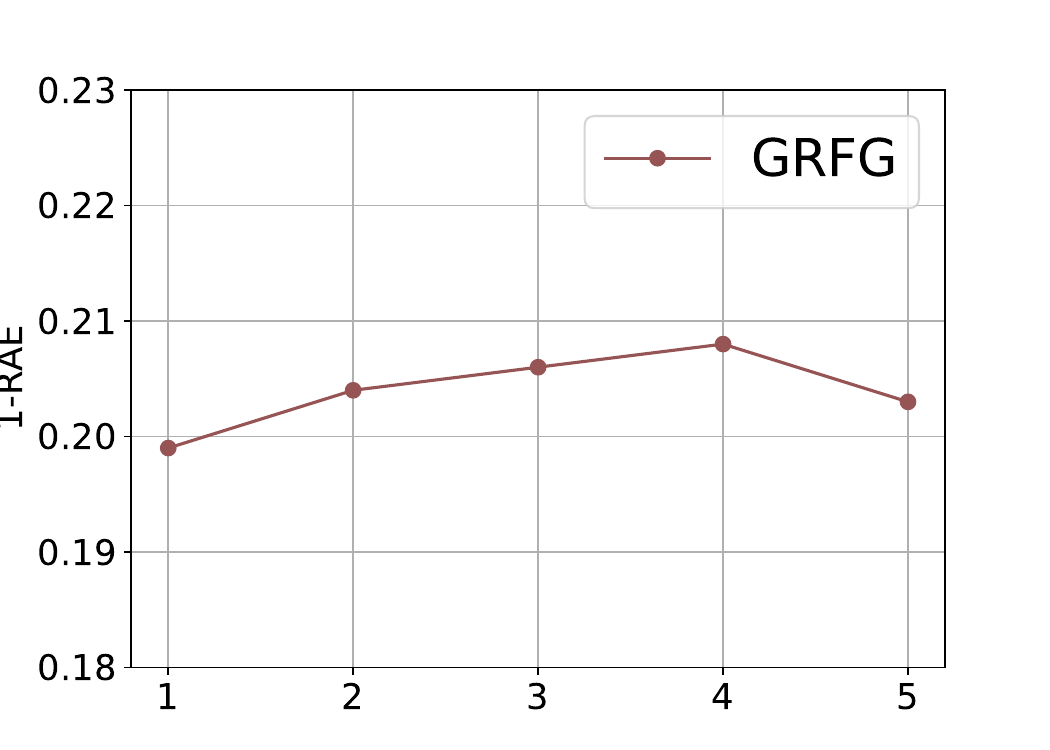}}
        \vspace{-5mm}
        \caption{Comparison of the most important features in the original descriptor space, and the \model\ generated space.}
        % \vspace{-5mm}
        \label{sen}
\end{figure}
To assess the sensitivity of \model\ to changes in hyperparameters, we perform a test focusing on the descriptor space size. We vary the size of the generated descriptor space by enlarging it from 1 to 5 times, indicating that the generated descriptor space ranges from 1 to 5 times the size of the original descriptor space.
Figure~\ref{sen} shows the results. It is evident that the optimal performance occurs when the size of the generated descriptor space is three times that of the original descriptor space. The potential reason is that when the generated space is large, it tends to encompass an abundance of redundant information. Conversely, when the generated space is too small, it may lack sufficient information. This experiment highlights the significance of controlling the appropriate size of the generated space.

\subsubsection{Case Study}
\begin{figure}[t]
\vspace{-1mm}
        \centering
        \subfigure[Original Descriptor Space]{\label{training}\includegraphics[width=0.23\textwidth]{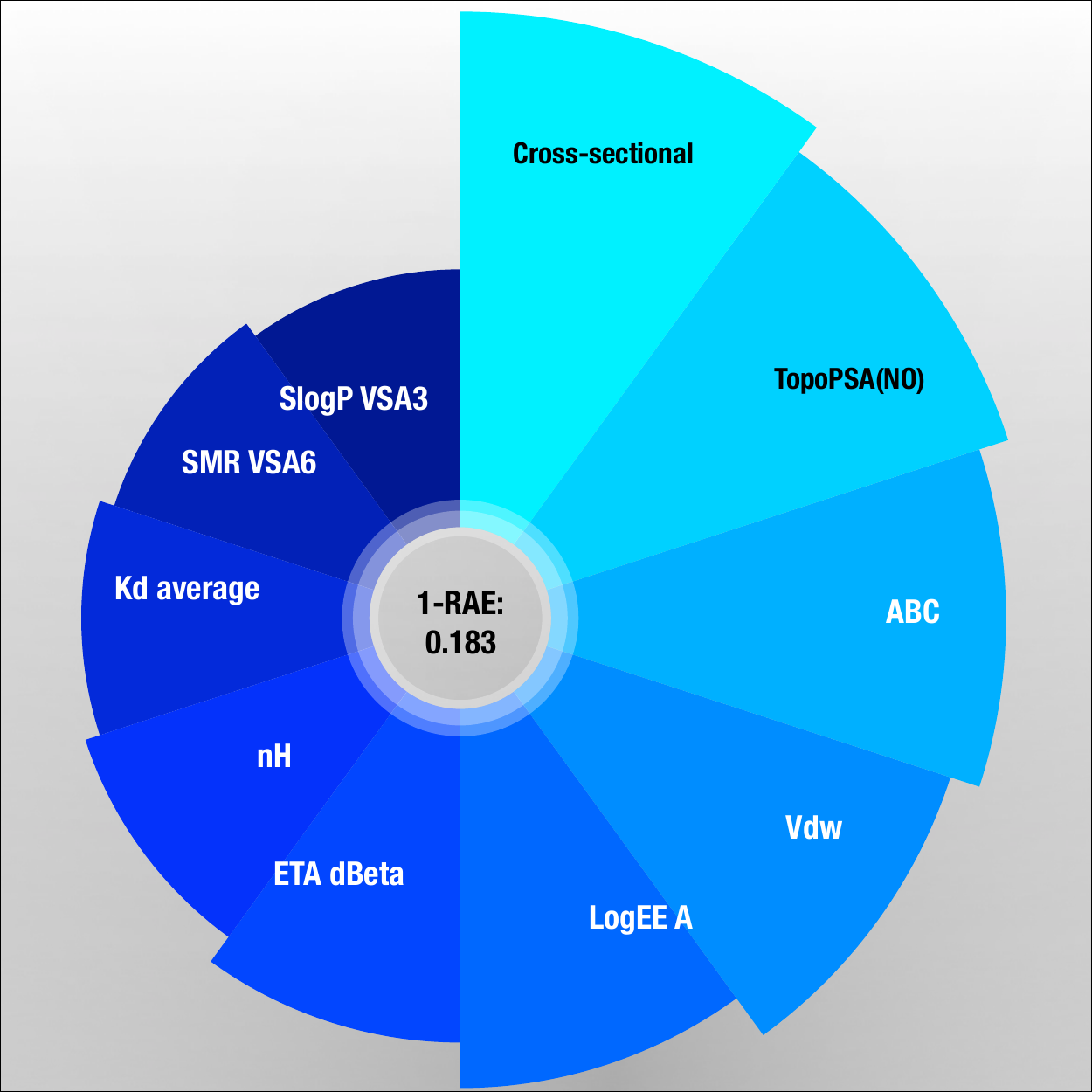}}
        \subfigure[\model\ Generated Space]{\label{generated}\includegraphics[width=0.23\textwidth]{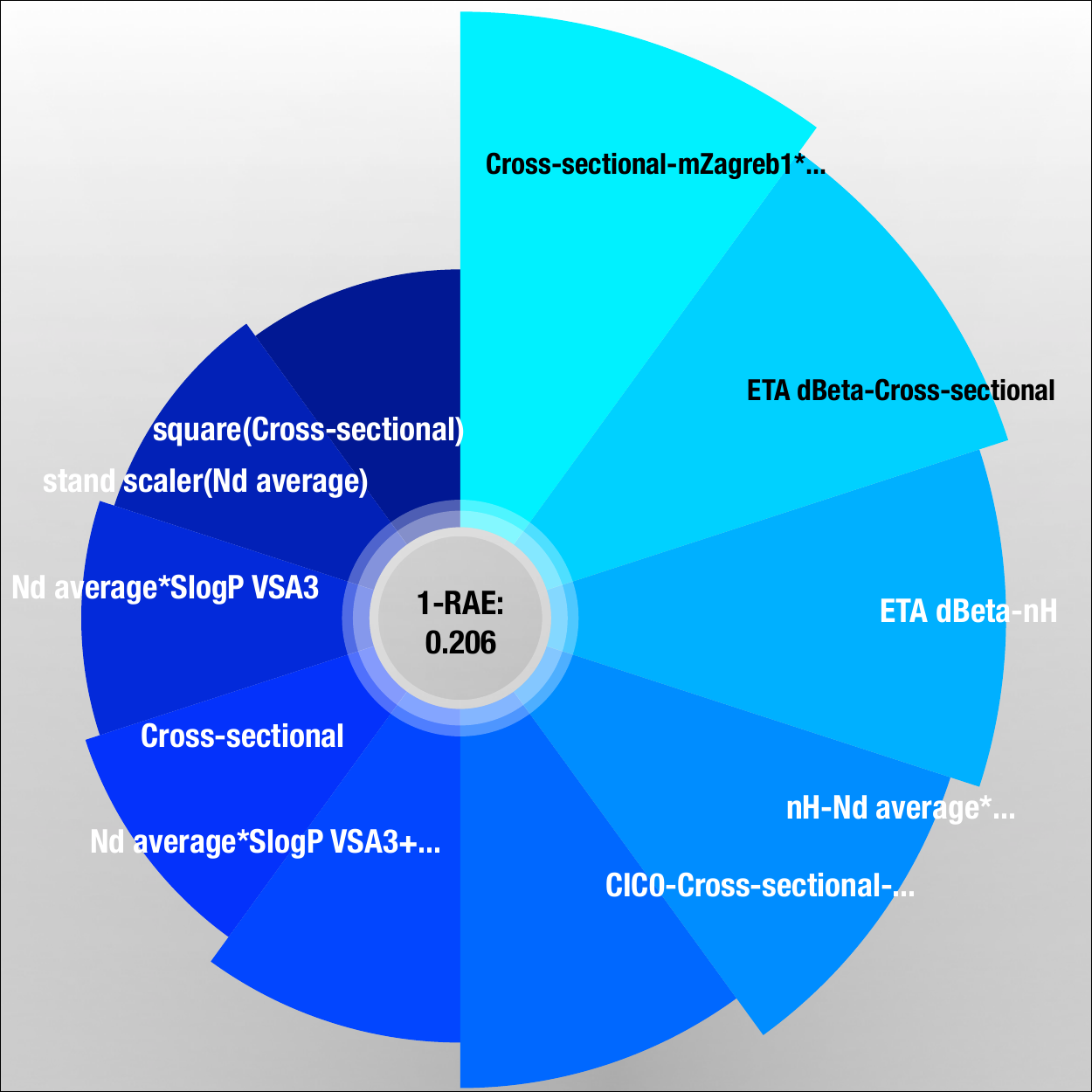}}
        \vspace{-4mm}
        \caption{Comparison of the most important features in the original descriptor space, and the \model\ generated space.}
        % \vspace{-7mm}
        \label{feature importance}
\end{figure}
To verify the interpretability and traceability of the descriptor space generated by \model\, we conduct this case study.
We pinpoint the top 10 most important features for polymer property performance prediction in both the original and reconstructed descriptor space to predict thermal conductivity, with random forest regression.
Figure~\ref{feature importance} indicates the result of case study. The labels associated with each pie chart indicate the corresponding feature names, and larger areas represent increased significance. We can observe that 90\% crucial descriptors in the generated space are generated by \model\ and they improve the accuracy of polymer property performance prediction by 12.6\%. This observation suggests that \model\ genuinely can understand the characteristics of both the descriptor set and machine learning models, leading to the creation of a more effective descriptor space. An additional noteworthy discovery is that ' [Cross-sectional] ' serves as the fundamental descriptor in the original descriptor set. However, \model\ goes a step further by generating additional mathematically composed descriptors utilizing '[Cross-sectional]'. This finding indicates that \model\ not only is able to identify significant descriptors but also generates more potent knowledge to enhance model performance. These composite descriptors enable material experts to trace their ancestral resources and formulate new analysis rules for evaluating the performance of polymer.
% \vspace{-3mm}
\section{Conclusion}
% \vspace{-1mm}
We introduce a framework called Group-wise Reinforcement Feature Generation (GRFG) designed for optimal and interpretable reconstruction of representation space of descriptors, with the goal of enhancing polymer property performance prediction performances. This framework integrates descriptor generation and selection in an iterative process to reconstruct a recognizable and size-controllable descriptor space through descriptor-crossing. Firstly, we break down the selection of crossing descriptors and operations into three Markov Decision Processes (MDPs) and devise a novel cascading agent structure for this purpose. Secondly, we propose two feature generation strategies based on cosine similarity and mutual information to address different generation scenarios following cascading selection. Thirdly, we advocate for a group-wise descriptor generation approach to efficiently create features and amplify rewards for cascading agents. To achieve this, we introduce a new descriptor clustering algorithm (M-Clustering) that generates robust descriptor groups from an information theory perspective. Extensive experiments demonstrate the effectiveness of GRFG in refining the descriptor space, showcasing competitive results against other baselines. Furthermore, GRFG provides traceable routes for descriptor generation, enhancing the explainability of the refined descriptor space. These results empirically demonstrate GRFG can generate an optimal and explainable representation space to improve the quality of polymer dataset and enhance the performance of prediction.

\section*{Acknowledgment}
This research was partially supported by the National Science Foundation (NSF) via the grant numbers: 2421864, 2421803, 2421865, and National academy of Engineering Grainger Foundation Frontiers of Engineering Grants.
\bibliographystyle{ACM-Reference-Format}
\balance
\bibliography{ref}
\end{document}